\pdfoutput=1

\documentclass[11pt]{article}

\usepackage[hyperref]{acl}

\usepackage{times}
\usepackage{latexsym}

\usepackage[T1]{fontenc}

\usepackage[utf8]{inputenc}

\usepackage{microtype}

\newcommand{\FirstBaseline}{\textsc{First}}
\newcommand{\SecondBaseline}{\textsc{Second}}
\newcommand{\RandomBaseline}{\textsc{Rand} (uni)}
\newcommand{\WeightedRandomBaseline}{\textsc{Rand} (dist)}

\newcommand{\CopyTitle}{Copy Title}
\newcommand{\SeqToSeq}{\textsc{S2S} + Ptr}
\newcommand{\HRED}{Hier \textsc{S2S} + Ptr}
\newcommand{\PLBARTFiltered}{PLBART (F)}

\newcommand{\TimeUnit}{time step}
\newcommand{\TitleStart}{$<$TITLE\_START$>$}
\newcommand{\UtteranceStart}{$<$UTTERANCE\_START$>$}

\newcommand{\HumanPLBARTFilt}{33.1}
\newcommand{\HumanTitle}{8.1}

\newcommand{\HumanHolistic}{62.4}

\newcommand{\FirstUtteranceOne}{$U_{1}$ (Lead 1)}
\newcommand{\FirstUtteranceThree}{$U_{1}$ (Lead 3)}
\newcommand{\LastUtteranceOne}{$U_{t_g}$ (Lead 1)}
\newcommand{\LastUtteranceThree}{$U_{t_g}$ (Lead 3)}
\newcommand{\ExSumm}{Supervised Extractive}
\newcommand{\RetTitleTitle}{Retrieval (Title-Title)}
\newcommand{\RetTitleDescription}{Retrieval (Title-Desc)}
\newcommand{\ProjectRetTitleTitle}{Project Retrieval (Title-Title)}
\newcommand{\ProjectRetTitleDescription}{Project Retrieval (Title-Desc)}

\newcommand{\Filtered}{Filtr.}
\newcommand{\BLEU}{\textbf{BLEU}}
\newcommand{\METEOR}{\textbf{METEOR}}
\newcommand{\ROUGEL}{\textbf{ROUGE}}
\newcommand{\Curated}{CS}
\newcommand{\CuratedCap}{CS}

\def\@fnsymbol#1{\ensuremath{\ifcase#1\or *\or \dagger\or \ddagger\or
   \mathsection\or \mathparagraph\or \|\or **\or \dagger\dagger
   \or \ddagger\ddagger \else\@ctrerr\fi}}
\newcommand{\ssymbol}[1]{^{\@fnsymbol{#1}}}

\usepackage{xcolor}

\usepackage{mdwlist}

\usepackage{multirow}
\usepackage{subcaption}
\usepackage{rotating}
\DeclareCaptionLabelFormat{noname}{#2}
\title{Learning to Describe Solutions for Bug Reports \\ Based on Developer Discussions}

\author{
Sheena Panthaplackel\textsuperscript{\rm 1},
\textbf{Junyi Jessy Li}\textsuperscript{\rm 2},
Milos Gligoric\textsuperscript{\rm 3},
\textbf{Raymond J. Mooney}\textsuperscript{\rm 1}\\
\textsuperscript{\rm 1}Department of Computer Science\\
\textsuperscript{\rm 2}Department of Linguistics\\
\textsuperscript{\rm 3}Department of Electrical and Computer Engineering\\
The University of Texas at Austin\\
\texttt{spantha@cs.utexas.edu, jessy@austin.utexas.edu}\\ \texttt{gligoric@utexas.edu, mooney@cs.utexas.edu}
}

\begin{document}
\maketitle

\begin{abstract}
When a software bug is reported, developers engage in a discussion to collaboratively resolve it. While the solution is likely formulated within the discussion, it is often buried in a large amount of text, making it difficult to comprehend and delaying its implementation. To expedite bug resolution, we propose generating a concise natural language description of the solution by synthesizing relevant content within the discussion, which encompasses both natural language and source code. We build a corpus for this task using a novel technique for obtaining noisy supervision from repository changes linked to bug reports, with which we establish benchmarks. We also design two systems for generating a description \textit{during} an ongoing discussion by classifying when sufficient context for performing the task emerges in real-time.
With automated and human evaluation, we find this task to form an ideal testbed for complex reasoning in long, bimodal dialogue context.
\end{abstract}

\section{Introduction}
Software bugs in open-source projects are reported through issue tracking systems like GitHub Issues.
When a bug is reported, a discussion is initiated among developers to collectively resolve it~\cite{NoyoriGood}.
The \textit{bug resolution} process is often strenuous and time-consuming, involving extended deliberations~\cite{LiuBugSum} among multiple participants~\cite{KavalerPerceived}, spanning long periods of time~\cite{KikasIssueDynamics}.
Although a solution often emerges within the discussion~\cite{AryaInfo}, this
can easily get lost in a large amount of text~\cite{LiuBugSum}. Wading through a long discussion to determine whether a solution has been suggested, comprehending it, and then implementing it can be daunting, especially for developers who are not closely following the discussion~\cite{AryaInfo,TanFirst}. Consequently, the resolution can be delayed.

\begin{figure}
\centering
\includegraphics[width=\columnwidth]{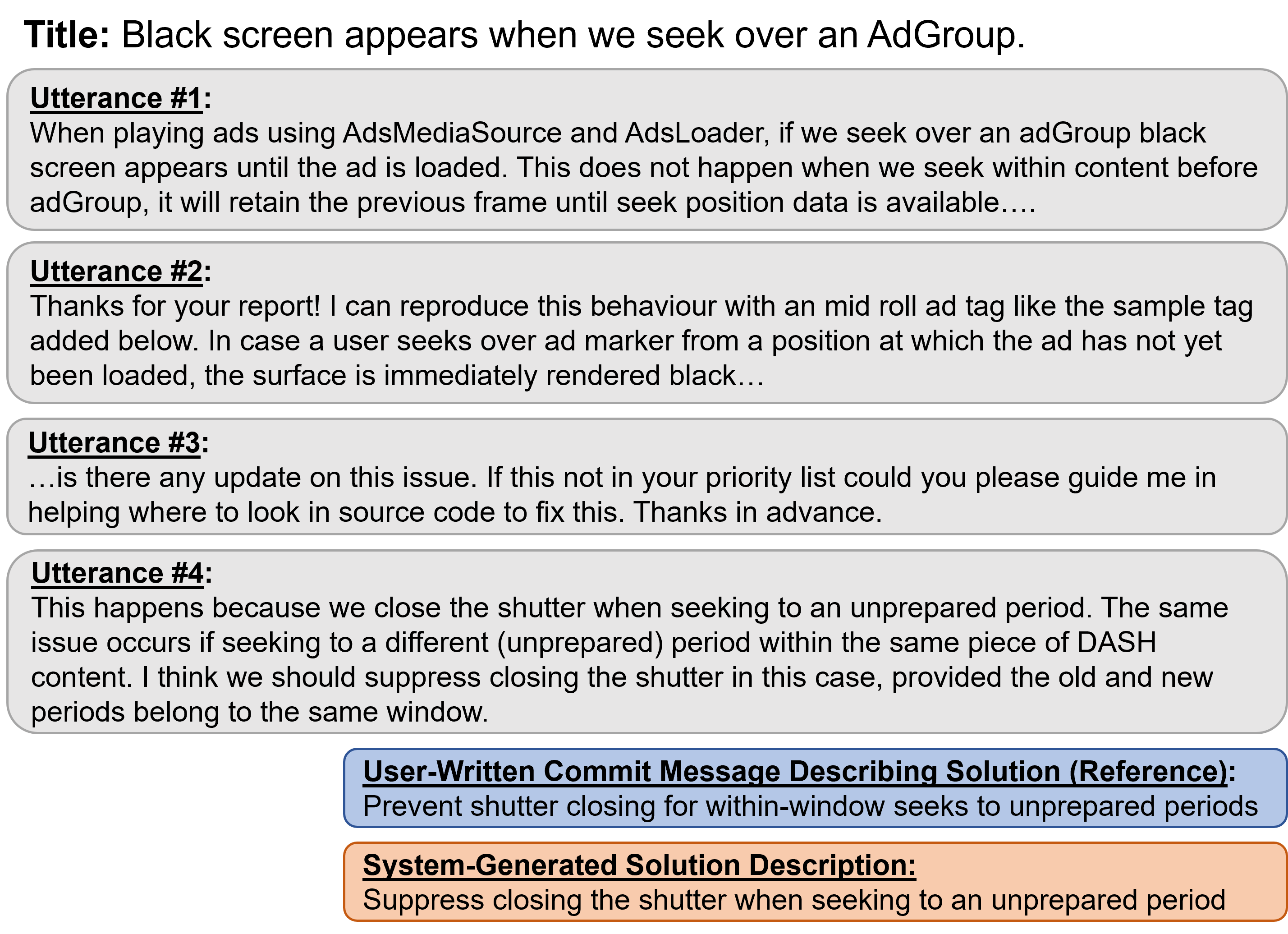}
\vspace{-10pt}
\caption[Caption for LOF]%
  {\small ExoPlayer bug report discussion with user-written and system-generated solution descriptions.}
\label{fig:main-ex}
\vspace{-10pt}
\end{figure}

As developers scan through the long discussion, it is desirable to have an automated system that guides them to more easily absorb information relevant towards implementing the solution. We propose automatically generating a concise natural language description of the solution by synthesizing relevant content as it emerges in the discussion.
For example, as the discussion in Figure~\ref{fig:main-ex} progresses, the cause of the bug is identified as the shutter getting closed ``when seeking to an unprepared period'' and a solution emerges: ``suppress closing the shutter in this case, provided the old and new periods belong to the same window.'' 
Our task aims to describe this solution:
\textit{Prevent shutter closing for within-window seeks to unprepared periods}.

To study this task, we build a corpus from bug report discussions on GitHub Issues. The changes made within the code base to resolve the bug are often linked to the bug report in the form of a commit or pull request. We develop a novel approach to obtain noisy supervision for the solution description from the associated commit message or pull request title which describe the bug-resolving changes in natural language. To control for noise, we apply filtering techniques. The dataset and code are publicly available for research use.\footnote{ \url{https://github.com/panthap2/describing-bug-report-solutions}}

With this data, we set benchmarks for generating solution descriptions, conditioned on the discussion. From the long context, a model must learn to tease out and condense information relevant to the solution. Handling long context is critical for tasks with dialogue as input, since the input grows rapidly with the number of interactions. Additionally, the context entails technical text, with natural language and source code often appearing in the same sentence~\cite{LiUnSupBugSum}. So, deducing information from the context to articulate a meaningful description requires complex reasoning. We explore generation models including transformer models~\cite{VaswaniTransformer} and PLBART~\cite{AhmadPLBART}, which was pretrained
on large quantities of code and technical text. We evaluate with automated metrics and human evaluation.

Furthermore, we investigate integrating our task into a real-time setting. 
An informative description can be generated only if there is sufficient context about the solution, so we must wait until this context is available. In Figure~\ref{fig:main-ex}, generation should be performed only after utterance \#4 is made in the discussion. Since the solution is not formulated until that point, there is insufficient context to reliably generate a description before then.
We design two methods for integrating a classifier that determines \textit{when} to generate with a generation model: (1) a \textit{pipelined} system with independently trained classification and generation models; (2) a \textit{joint} system that is simultaneously trained for both tasks.

By monitoring progress and later chiming into the discussion with a solution description, this combined system lays the groundwork for future work on developing an intelligent dialogue agent which participates in discussions to facilitate more efficient bug resolution. While there is growing interest in building tools to support development activities such as code summarization~\cite{iyer-etal-2016-summarizing, ahmad-etal-2020-transformer}, comment updating~\cite{panthaplackel-etal-2020-learning}, and commit message generation~\cite{loyola-etal-2017-neural}, dialogue systems have been largely understudied in this domain. We consider our work as a step towards building more dialogue-based AI tools for software development.

\section{Problem Setting}
As shown in Figure~\ref{fig:main-ex}, when a user reports a bug, they state the problem in the \textit{title} (e.g., ``Black screen appears when we seek over an AdGroup") and initiate a discussion by making the first \textit{utterance} ($U_1$), which usually elaborates on the problem.
Other participants join the discussion at later time steps through utterances ($U_2...U_T$), where $T$ is the total number of utterances. Throughout the discussion, developers discuss various aspects of the bug, including a potential solution~\cite{AryaInfo}. We propose the task of generating a concise description of the solution (e.g., ``Prevent shutter closing for within-window seeks to unprepared periods") by synthesizing relevant content within the title and sequence of utterances ($U_1, U_2...$).

\section{Data}

\begin{table*}[t]
\begin{center}
\small
\begin{tabular}{lllll}
\hline
 & \bf Train & \bf Valid & \bf Test & \bf Total \\
 \hline
  Projects & 395 (330) & 145 (111) & 134 (104) & 412 (344) \\
 Examples & 9,862 (4,664) & 1,232 (599) & 1,234 (593) & 12,328 (5,856) \\
 \hspace{3mm}\# Commit messages & 4,520 (2,355) & 410 (234) & 386 (189) & 5,316 (2,778) \\
 \hspace{3mm}\# PR titles & 5,342 (2,309) & 822 (365) & 848 (404) & 7,012 (3,078) \\
 Avg $T$ & 3.9 (4.5) & 3.8 (4.4) & 4.0 (4.4) & 3.9 (4.5) \\
 Avg $t_g$ & 2.9 (3.4) & 2.9 (3.4) & 3.2 (3.6) & 2.9 (3.4) \\
 Avg utterance length (\#tokens) & 68.4 (75.6) & 74.8 (84.3) & 70.2 (75.7) & 69.2 (76.5) \\
 Avg title length (\#tokens) & 10.6 (10.6) & 11.2 (11.0) & 11.5 (11.3) & 10.7 (10.7) \\
 Avg description length (\#tokens) & 9.1 (10.5) & 8.9 (9.9) & 9.1 (10.1) & 9.1 (10.4) \\

\hline

\end{tabular}
\end{center}
\vspace{-10pt}
\caption{\small \label{table:data-table}Data statistics. In parentheses, we show metrics computed on the filtered subset.}
\vspace{-10pt}
\end{table*}

\label{sec:data}
Following prior work on other tasks~\cite{KavalerPerceived, PanchellaWont}, we mine issue reports corresponding to open-source Java projects from GitHub Issues.
Issue reports can entail feature requests as well as bug reports. In this work, we focus on the latter. We identify bug reports by searching for ``bug" in the labels assigned to a report and by using a heuristic for identifying bug-related commits~\cite{KarampatsisBugs}.

\subsection{Data Collection}
\label{sub:data_collection}
A bug report is organized as an event timeline, recording activity from when it is opened to when it is closed. From comments that are posted on this timeline, we extract utterances which form the \textit{discussion} corresponding to a bug report, ordered based on their timestamps. 
We specifically consider bug reports that resulted in code (or documentation) fixes~\cite{NguyenCommit}. These changes are made through \textit{commits} and \textit{pull requests}, which also appear on the timeline. Changes made in a commit or pull request are described using natural language, in the corresponding commit message~\cite{loyola-etal-2017-neural, XuCommit} or pull request title~\cite{KononenkoShopify, Zhao2019ImprovingTP}. In practice, commit messages and pull request titles are written after code changes. However, like contemporary work~\cite{MultiModalChakraborty}, we treat them as a proxy for solution descriptions to drive bug-resolving code changes.

Furthermore, we extract the position of a commit or pull request on the timeline, relative to the utterances in the discussion.
We consider this as the point at which a developer acquired enough information about the solution to implement the necessary changes and describe these changes with the corresponding commit message or pull request title. So, if the implementation is done immediately after $U_g$ on the timeline, then we take this position $t_g$ as the ``gold" \TimeUnit{} for when sufficient context becomes available to generate an informative description of the solution. This leads to examples of the form \textit{(Title, $U_{1}...U_{T}$, $t_{g}$, description)}.

We disregard issues with multiple commit messages/PR titles, so there is at most one example per issue. This is because
the reason for needing multiple sets of changes is not clear (e.g., the solution could be implemented in parts or the first solution may have been incorrect and it is later corrected).\footnote{However, since such examples could be useful for future work, they are available in the data we release.}

\subsection{Handling Noise}
\label{sec:data:noise}

Due to significant noise in large online code bases like GitHub and StackOverflow, automatically extracted data from these sources is typically filtered both for more effective supervision and for more accurate evaluation~\cite{panthaplackel2020associating, allamanis2016convolutional,Hu2018DeepCC,Fernandes2018StructuredNS, iyer-etal-2016-summarizing,yao2018staqc, yin2018mining}. Upon studying the data, we also deemed filtering to be necessary. 
First, we apply simple heuristics to reduce noise, which we discuss in more detail in Appendix~\ref{sec:data-cleaning}.
From this, we obtain the examples that are primarily used for training and evaluation in this work, which we refer to as the \textit{full dataset}.
Next, we identify three sources of noise that 
 are more difficult to control with simple heuristics
 and use techniques described below to quantify them and build a \textit{filtered subset} of the full dataset that is less noisy.  This subset is used for more detailed analysis of the models that are discussed in the paper, and we find that training on this subset leads to improved performance (\S\ref{sec:HowHumanEval}).

\noindent\textbf{Generic descriptions}: Commit messages and pull request titles are sometimes generic (e.g., \textit{``fix issue.''})~\cite{EtemadiCommit}.
To limit such cases, we compute normalized inverse word frequency (NIWF), which is used in prior work to quantify specificity~\cite{zhang-etal-2018-learning-control}. The filter excludes 1,658
examples in which the reference description's NIWF score is below 0.116 (10th percentile computed from the training data). 

\noindent\textbf{Uninformative descriptions}: Instead of describing the solution, the commit message or pull request title 
sometimes
essentially re-states the problem (which is usually mentioned in the title of the bug report). To control for this, we compute the percentage of unique, non-stopword tokens in the reference description which also appear in the title. The filtered subset excludes 3,552 additional examples in which this percentage is 50\% or more.

\noindent\textbf{Discussions without sufficient context}: While enough context is available to a developer to implement a solution at $t_g$, this context may not always be available in the discussion and could instead be from their technical expertise or external resources. 
For instance, in the discussion in the footnote\footnote{\url{https://github.com/prestodb/presto/issues/14567}}, only a stack trace and personal exchanges between developers are present. From the utterance before the PR, ``Or PM me the query that failed" suggests that an offline conversation occurred. Since relevant content is not available in such cases, it is unreasonable to expect to generate an informative description. We try to identify such examples
with an approach~\cite{NallapatiSumma} for greedily
constructing an extractive summary based on a reference abstractive summary.
The filtered subset excludes 1,262 more examples for which a summary could not be constructed (i.e., there is no relevant sentence that is extracted from the context).
After applying all three filters, we have 5,856 examples.

\subsection{Preprocessing}
Since text in this domain can contain code tokens, we \textit{subtokenize} (e.g., snake\_case $\rightarrow$ snake case, camelCase $\rightarrow$ camel case) in the title, utterances, and description. We retain inlined code (on average 5.7 tokens/utterance); however, we remove code blocks and embedded code snippets (with markdown tags), as done in prior work~\cite{tabassum-etal-2020-code, AhmadPLBART}. Capturing meaning from large bodies of code often requires reasoning with respect to the abstract syntax tree~\cite{AlonCode2Seq} and data and control flow graphs~\cite{AllamanisGraph}. However, markdown tags are not always used to identify code~\cite{tabassum-etal-2020-code}, and consequently, we observe some instances of larger code blocks within utterances that cannot be easily removed.
We do not use source code files within a project's repository and leave it to future work to incorporate large bodies of code.
We discard URLs and mentions of GitHub usernames from utterances. From the description, we remove references to issue and pull request numbers.

\subsection{Partitioning}
\label{sec:data-partitioning}
The dataset spans bug reports from April 2011 - July 2020. We partition based on the timestamp of the commit or pull request associated with a given example. Namely, we require all timestamps in the training set to precede those in the validation set and those in the validation set to precede those in the test set.
Partitioning with respect to time ensures that we are not using models trained on future data to make predictions in the present, more closely resembling the real-world scenario~\cite{NieTime}.
Dataset statistics are shown in Table~\ref{table:data-table}.

\section{Generating Solution Descriptions}

\begin{table}
\begin{center}
\small
\begin{tabular}{llcccc}
\hline

& & \bf 1 & \bf 2 & \bf 3 & \bf 4 \\
\hline
\multirow{3}{*}{Full}
& Title & 73.0 & 88.9 & 94.0 & 96.1\\
& $U_1...U_{t_g}$ & 54.7 & 87.6 & 95.0 & 97.6 \\
& Title + $U_1...U_{t_g}$ & 47.9 & 82.0 & 91.2 & 94.8 \\
\hline
\multirow{3}{*}{\Filtered{}}
& Title & 82.3 & 95.6 & 98.4 & 99.4\\
& $U_1...U_{t_g}$ & 49.9 & 87.4 & 95.1 & 97.8\\
& Title + $U_1...U_{t_g}$ & 47.5 & 86.0 & 94.5 & 97.5\\
\hline

\end{tabular}
\end{center}
\vspace{-10pt}
\caption{\small \label{table:novel_ngrams}Percent of novel unigrams, bigrams, trigrams, and 4-grams in the reference description, with respect to the title, $U_1...U_{t_g}$, and title + $U_1...U_{t_g}$. High percentages show that generating solutions is an abstractive task.}
\vspace{-10pt}
\end{table}

We first generate informative solution descriptions in a static setting, in which we leverage the oracle context from the discussion (i.e., the title and  $U_1...U_{t_g}$). From Table~\ref{table:data-table}, the average length of a single utterance is $\sim$70 tokens while the average description length is only $\sim$9 tokens. Therefore, this task requires not only effectively selecting content about the solution from the long context (which could span multiple utterances) but also synthesizing this content to produce a concise description. Following \citet{see-etal-2017-get}, we compute the percent of novel n-grams in the reference description with respect to the input context in Table~\ref{table:novel_ngrams}. The high percentages underline the need for an \textit{abstractive} approach, rather than an \textit{extractive} one which generates a description by merely copying over utterances or sentences %
within the discussion.\footnote{We observe very low performance with extractive approaches, as shown in Appendix~\ref{app:gen-baselines}.} Furthermore, this task requires complex, bimodal reasoning over the discussion, encompassing both natural language and source code. 

\subsection{Models}
\label{sec:gen-models}
We benchmark various models for this task.
To represent the input in neural models, we insert \TitleStart{} before the title and \UtteranceStart{} before each utterance.

\noindent\textbf{\CopyTitle}: Though the bug report title usually only states a problem, we observe that it sometimes also puts forth a possible solution, so
we evaluate how well it can serve as a solution description.

\noindent\textbf{\SeqToSeq}: We consider a transformer encoder-decoder model~\cite{VaswaniTransformer} in which we flatten the context into a single input sequence.
Generating the output often requires incorporating project-specific out-of-vocabulary tokens from the input, so we support copying with a pointer generator network~\cite{VinyalsPointer}.

\noindent\textbf{\HRED}: Inspired
by hierarchical approaches for dialogue response generation~\cite{SerbanHRED}, we consider a hierarchical variant of the \SeqToSeq{} model with two separate encoders: one for representing an individual utterance, and one for representing the whole discussion. We provide implementation details in Appendix~\ref{app:hred-details}.

\noindent\textbf{PLBART}:
\citet{AhmadPLBART} proposed PLBART, which is pretrained on a large amount of code from GitHub and software-related natural language from StackOverflow, using BART-like~\cite{lewis-etal-2020-bart} training objectives.
With fine-tuning, PLBART achieves state-of-the-art performance on many program and language understanding tasks.
We fine-tune PLBART on our training set and evaluate its ability to comprehend bug report discussions and generate descriptions of solutions.\footnote{We focus on PLBART rather than vanilla BART because it achieves higher performance, as shown in Appendix~\ref{app:bart}.}
Note that PLBART has a 1024 token limit. We use left truncation to keep the most recent content.

\noindent\textbf{\PLBARTFiltered}: Since PLBART is pretrained on a large amount of data, we can afford to reduce the fine-tuning data. So we fine-tune on only the filtered subset of the training set (cf. \S\ref{sec:data:noise}), to investigate whether fine-tuning on this ``less noisy" sample can lead to improved performance.

\begin{table}[t]
\begin{center}
\small
\begin{tabular}{lllll}
\hline
& \bf Model & \BLEU & \METEOR & \ROUGEL \\
\hline
\multirow{5}{*}{\rotatebox{90}{Full}}
& \CopyTitle{} & 14.4$\ssymbol{6}$ & 13.1 &  24.4$\ssymbol{4}$\\
& \SeqToSeq{} & 12.6 & 9.8 & 25.0$\ssymbol{3}$ \\
& \HRED{} & 12.4 & 9.6 & 24.1$\ssymbol{4}$ \\
& PLBART & \bf 16.6 & \bf 14.5 & \bf 28.3 \\
& \PLBARTFiltered{} & 14.2$\ssymbol{6}$ & 12.3 & 25.1$\ssymbol{3}$\\
\hline
\multirow{5}{*}{\rotatebox{90}{\Filtered{}}}
& \CopyTitle{} & 10.0$\ssymbol{1}$$\ssymbol{2}$ & 8.3 & 16.6\\
& \SeqToSeq{} & 10.2$\ssymbol{1}$ & 7.5 & 20.1\\
& \HRED{} & 9.9$\ssymbol{2}$ & 7.4 & 19.6\\
& PLBART & \bf 12.3$\ssymbol{3}$ & 9.9 & 21.1  \\
& \PLBARTFiltered{} & \bf 12.3$\ssymbol{3}$ & \bf 10.2 & \bf 21.9\\

\hline

\end{tabular}
\end{center}
\vspace{-10pt}
\caption{\small \label{table:gen-results-table}Automated metrics. \SeqToSeq{} and \HRED{} scores are averaged across 3 trials. Differences that are \textit{not} statistically significant are indicated with matching symbols.}
\vspace{-10pt}
\end{table}

\subsection{Results: Automated Metrics}
\label{sec:HowAutoEval}

We use text generation metrics, BLEU-4~\cite{papineni2002bleu}, METEOR~\cite{BanerjeeEtAL2005}, and ROUGE-L~\cite{lin-2004-rouge}. We compute statistical significance with bootstrap tests~\cite{berg-kirkpatrick-etal-2012-empirical} with $p < 0.05$. Results are in Table~\ref{table:gen-results-table}. On the full test set, PLBART significantly outperforms other models, demonstrating the value of pretraining on large amounts of data. \PLBARTFiltered{}
underperforms PLBART on the full test set. On the filtered subset, it either beats or matches PLBART.

Performance drops across models between the full and filtered test sets. The relatively high performance of the naive \CopyTitle{} baseline shows that simply copying or rephrasing the title performs well in many cases, particularly for the full test. The filtered subset is designed to remove uninformative reference descriptions that merely re-state the problem, as illustrated in Table~\ref{table:novel_ngrams} with filtered reference descriptions having higher percentages of novel n-grams, with respect to the title. Nonetheless, keywords relevant to the solution are often also in the title, so the \CopyTitle{} baseline still achieves reasonable scores on the filtered subset. Although automated metrics provide some signal, they emphasize syntactic similarity over semantic similarity. So, we conduct human evaluation.

\subsection{Results: Human Evaluation}
\label{sec:HowHumanEval}
 Evaluators first read the title and discussion ($U_{1}...U_{t_{g}}$).
 For each example, they are shown predictions from the 5 models discussed in Section~\ref{sec:gen-models}. From these, they must select one or more that are most informative towards resolving the bug. 
 If all candidates are uninformative, they select a separate option: ``All candidates are poor." There is also another option to indicate that there is insufficient context about the solution (\S\ref{sec:data:noise}), making it difficult to evaluate candidate descriptions.
 They also write a rationale for their response.
 


\begin{table}[t]
\begin{center}
\small
\begin{tabular}{lll}
\hline
\bf Model & \bf Full & \bf Filtered \\
\hline
\CopyTitle{} & 8.1 & 6.0 \\
\SeqToSeq{} & 1.3$\ssymbol{1}$ & 1.2$\ssymbol{2}$ \\
\HRED{} & 1.3$\ssymbol{1}$ & 1.2$\ssymbol{2}$ \\
PLBART & 11.9 & 10.5 \\
\PLBARTFiltered{} & \bf 33.1$\ssymbol{3}$ & \bf 39.5 \\
\hline
All Poor & 20.0 & 22.1 \\
Insufficient Context & 31.9$\ssymbol{3}$ & 25.6 \\

\hline
\end{tabular}
\end{center}
\vspace{-10pt}
\caption{\small \label{table:human-eval}Human evaluation results: Percent of annotations for which users selected predictions made by each model. This entails 160 annotations for the full test set, 86 of which correspond to examples in our filtered subset. Differences that are \textit{not} significant are indicated with matching superscripts.}
\vspace{-10pt}
\end{table}

Since annotation requires not only technical expertise, but also high cognitive load and time commitment, it is hard to perform human evaluation on  a large number of examples with multiple judgments per example. Similar to \citet{iyer-etal-2016-summarizing}, we resort to having each example annotated by one user to obtain more examples. 
We recruited 8 graduate students  with 3+ years of programming experience and familiarity with Java.
They are not active contributors, so they will likely select
the option of insufficient context more often than if they had a deeper understanding of the various software projects.
However, it is difficult to conduct a user study at a similar scale with contributors. Nonetheless, there are developers aiming to become first-time contributors for a particular project~\cite{TanFirst}. Our study better aligns with this use case.

Each user annotated 20 examples, leading to annotations for 160 unique examples in the full test set. 
In Table~\ref{table:human-eval}, we show that \PLBARTFiltered{} substantially outperforms all other models, with users selecting its output  \HumanPLBARTFilt{}\% of the time. Even though the title typically only states a problem, users selected it  \HumanTitle{}\% of the time. From rationales
that users were asked to write, we found that there were cases in which the title not only posed the problem but also offered a solution. Users rarely preferred the output of \SeqToSeq{} and \HRED{} as they usually just rephrased the problem. PLBART also appears to be re-stating the problem in many cases; however, less often than other models.

Though we see similar trends across the full test set and the filtered subset, all models except \PLBARTFiltered{} tend to perform worse on the filtered subset, as previously observed on automated metrics. Also, the average number of cases with insufficient context is lower for the filtered subset, confirming that we are able to reduce such cases through filtering. We find the results on the filtered data to align better with human judgment. By fine-tuning on the filtered training set, \PLBARTFiltered{} learns to pick out important information from within the context and generate descriptions which reflect the solution rather than the problem.

\subsection{Analysis}
\label{sec:curated_analysis}
Of the 160 annotated examples, users found 109 to have sufficient context about the solution. We consider this the \textit{context-sufficient subset (\Curated{})}, which we will release for future research.
To analyze how models exploit the provided context, we measure the percent of n-grams in the prediction which overlap with the title as well as $U_1...U_{t_g}$  (excluding n-grams already in the title) in Table~\ref{table:curated_ngrams}.
\PLBARTFiltered{}'s predictions tend to have less n-gram overlap with the title and more overlap with the utterances.
This suggests that this model predicts fewer uninformative descriptions which merely re-state the problem mentioned in the title and instead focuses on other content from the utterances.

\begin{table}[t]
\begin{center}
\small
\begin{tabular}{lccccc}
\hline
& \multicolumn{2}{c}{\bf Title $\downarrow$} && \multicolumn{2}{c}{\bf $U_1...U_{t_g}$ only $\uparrow$}  \\
\cline{2-3}
\cline{5-6}
\bf  Model &  \bf 1 &  \bf 2 &&  \bf 1 &  \bf 2  \\
\hline
\CopyTitle{} & 100.0 & 100.0  && 0.0 & 0.0 \\
\SeqToSeq{} & 64.8 & 37.1  && 31.6 & 25.3\\
\HRED{} & 60.3 & 34.2  && 38.7 & 26.1 \\
PLBART & 80.8 & 77.7 && 31.0 & 41.4 \\
\PLBARTFiltered{} & 36.9 & 28.4  &&  \bf 52.8 &  \bf 42.3\\
Reference &  \bf 32.7 & \bf 22.2  && 38.8 & 25.4 \\

\hline

\end{tabular}
\end{center}
\vspace{-10pt}
\caption{\small \label{table:curated_ngrams}Percent of unigrams and bigrams in the prediction (or reference) which appear in the title and in $U_1..U_{t_g}$ only (excluding the title), on the \Curated{} subset. }
\end{table}

\begin{table}[t]
\begin{center}
\small
\begin{tabular}{p{2cm}p{4.6cm}}
\hline
\textbf{Model} & \textbf{Prediction} \\
\hline
\CopyTitle{} & black screen appears when we seek over an ad group . \\
\hline
\SeqToSeq{} & fix black ads \\
\hline
\HRED{} & fix seeking in ad tag \\
\hline
PLBART & suppress closing shutter when seeking over an ad group \\
\hline
\PLBARTFiltered{} & suppress closing the shutter when seeking to an unprepared period \\
\hline

Reference & prevent shutter closing for within - window seeks to unprepared periods \\
\hline
\end{tabular}
\vspace{-8pt}
\caption{\small \label{table:all-preds} Model outputs for the example shown in Figure~\ref{fig:main-ex}.}
\vspace{-15pt}
\end{center}
\end{table}

In Table~\ref{table:all-preds}, we show model outputs for the example in Figure~\ref{fig:main-ex}. SeqToSeq{} and \HRED{} essentially rephrase aspects of the problem, which are described in the title. Both PLBART and \PLBARTFiltered{} capture the solution, with \PLBARTFiltered{} providing more information. 
When there is sufficient context, \HumanHolistic{}\% of the time, either PLBART or \PLBARTFiltered{} generates output that is informative towards bug resolution. While this demonstrates that fine-tuning this large, pretrained model on our data can be useful in supporting bug resolution in on-line discussions to some extent, it also shows that there is opportunity for improvement.

\begin{table*}[h]
\begin{center}
\small
\begin{tabular}{lp{5.6cm}p{4cm}p{4cm}}
\hline

& \bf Title & \textbf{PLBART (F)}  & \textbf{Reference}\\
\hline
(1) & Issue with dex: OIDC server is not available at the 'quarkus.oidc.auth-server-url' URL & fix trailing slash in auth - server url & strip trailing forward slash from oidc url \\
\hline
(2) & InvalidDataTypeException: UDATA contains value larger than Integer.MAX\_VALUE DDR issue decoding lookswitch & fix bug in byte code dumper when tableswitch instruction precedes tableswitch instruction &  fix interpretation of switch instructions in byte code dumper \\
\hline
(3) & Worldmap viewport changes when switching between dashboard pages & don ' t refresh widget grid when worldmap loses viewport & define key prop for map visualization to update map on dimension change \\
\hline
(4) & Workaround comments exist in opengrok-indexer/pom.xml file while the related issues are already fixed. & fix jflex - de / jflex \# 705 ( comment ) & use jflex 1.8.2 \\
\hline
(5) & Why subscribe with single action for onNext design to crush if error happened? & 1 . x : fix subscription . subscribe ( ) to return observable . empty ( ) 2 . x : fix subscription . subscribe ( ) to return observable . empty ( ) & fixed sonar findings \\

\hline 
\end{tabular}
\end{center}
\vspace{-10pt}
\caption{\small \label{table:error-analysis} Output of \PLBARTFiltered{} for a sample of examples in the test set. Derived from: \url{https://github.com/quarkusio/quarkus/issues/10227}, \url{https://github.com/eclipse-openj9/openj9/issues/9294}, \url{https://github.com/Graylog2/graylog2-server/issues/7997}, \url{https://github.com/oracle/opengrok/issues/3172}, \url{https://github.com/ReactiveX/RxJava/issues/637}.
}
\vspace{-10pt}
\end{table*}

We manually inspected \PLBARTFiltered{}'s outputs and associated user rationales.  We observe that the model tends  to perform better when the solution is clearly stated in 1-3 consecutive sentences (Table~\ref{table:error-analysis} (1) and (2)). When more complex synthesis is needed, it sometimes stitches together tokens from the input incorrectly (Table~\ref{table:error-analysis} (3)). Next, although the model picks up on information in the context, sometimes, it draws content from an elaboration of the problem from within the discussion rather than a formulation of the solution (Table~\ref{table:error-analysis} (4)). This demonstrates that it still struggles to disentangle content relevant to the solution from that about the problem. It also sometimes struggles to generate meaningful output when in-lined code is present, highlighting the challenge in bimodal reasoning about code and natural language (Table~\ref{table:error-analysis} (5)).

\section{Supporting Real-Time Generation}

Generating an informative description requires sufficient context about the solution being available in the discussion. In a real-time setting, this context is likely not immediately available but rather emerges as the discussion progresses, and we must wait until it becomes available
to generate a solution description. However, the \TimeUnit{} at which it becomes available ($t_g$) is not known beforehand, so we must instead predict it ($t_p$) in order to perform generation \textit{during} ongoing discussions. For this, we consider classifying whether sufficient context is available upon each new utterance. In Figure~\ref{fig:main-ex}, the solution is formulated in $U_4$, so the correct behavior is to predict the negative label at $t=1,2,3$ and the positive label at $t=4$. Once the positive label is predicted at $t_p$\footnote{Classifications are not made at $t > t_p$. We leave generating at multiple time steps for future work.}, the description is generated, conditioned on the title and $U_1...U_{t_p}$. We develop two systems for integrating classification with a generation model: \textit{pipelined} and \textit{joint trained}.

\subsection{Pipelined System}
\label{sec:pipeline}
We design an independent classifier built on PLBART's encoder.
When a new utterance $U_t$ is made in the discussion, we encode the context so far (the title and all utterances up to and including $U_t$).
We take the final hidden state, $e_t$, as the context representation at $t$, which we feed $e_t$ through a 3-layer classification head
and apply softmax to classify whether or not sufficient context is available.
We train to minimize cross entropy loss.
At test time, we use the already trained \PLBARTFiltered{} model to generate a solution description with context available at $t_p$.

\subsection{Joint System}
We initialize an encoder-decoder model from PLBART with an additional classification head (\S\ref{sec:pipeline}).
The encoder is shared among the two tasks.
When classifying whether sufficient context about the solution is available, there is likely specific solution-related content that contributes to predicting the positive label. So, classification may enhance encoder representations, improving content selection for generating solution descriptions. 

Furthermore, having sufficient context correlates with whether it can be used to generate an \textit{informative} description. So, the informativeness of a description that can be generated with the available context can provide signal for classifying whether that context is sufficient. Additionally, if sufficient context was not previously available at $t-1$ but becomes available at $t$, we expect an improvement in the informativeness of the descriptions generated at the two time steps. We represent these descriptions with the final decoder states at the two time steps, $d_{t-1}$ and $d_{t}$. We concatenate $e_t$, $d_{t-1}$, and $d_t$ to form the input into the classification head.
For training loss, we sum the generation and classification losses across time steps $t_1...t_{g}$. Sufficient context for generation may not be available at $t < t_g$, so we mask  generation loss for earlier time steps.

\begin{table*}[t]
\begin{center}
\small
\begin{tabular}{lllllll}
\hline
&  & $\mathbf{t_p \leq t_g }$ & $\mathbf{t_g - t_p}$  & \bf BLEU & \bf METEOR & \bf ROUGE \\
\hline
\multirow{4}{*}{Pipelined} &\multirow{2}{*}{Full}
& @$t_p$ 
& 1.69 & 14.3$\ssymbol{3}$ & 12.4$\ssymbol{4}$ & 25.1$\ssymbol{5}$ \\
&& @$t_g$ 
& \bf - & \bf 14.4$\ssymbol{3}$ & \bf 12.5$\ssymbol{4}$ & \bf 25.3$\ssymbol{5}$ \\
\cline{2-7}
&\multirow{2}{*}{\Filtered{}}
&  @$t_p$ 
& 1.85 & 12.5$\ssymbol{1}$ & 10.1 & 21.7 \\
&& @$t_g$ 
& \bf - & \bf 12.6$\ssymbol{1}$ & \bf 10.5 &  \bf 22.3
 \\

\hline

\multirow{4}{*}{Joint} 
& \multirow{2}{*}{Full}
& @$t_p$ 
& 1.81 & 13.1 & 11.4 & 22.4$\ssymbol{2}$ \\
&& @$t_g$ 
& \bf - & \bf 13.2 & \bf 11.7 & \bf 22.5$\ssymbol{2}$
 \\
\cline{2-7}
&\multirow{2}{*}{\Filtered{}}
&  @$t_p$ 
& 1.97 & 11.7
 & 9.5
 & 19.3
 \\
&& @$t_g$ 
& - & \bf 11.9 & \bf 9.9 & \bf 19.7
 \\
 \hline

\end{tabular}
\end{center}
\vspace{-10pt}
\caption{\small \label{table:combined-automatic}Automated metrics for combined systems when $t_p \leq t_g$. We compare the generated description @$t_p$ with that if the system had generated @$t_g$. Differences that are \textit{not} statistically significant are indicated with matching superscripts.}
\end{table*}

\begin{table*}[t]
\begin{center}
\small
\begin{tabular}{llllll}
\hline
&  & $\mathbf{t_g - t_p}$  & \bf BLEU & \bf METEOR & \bf ROUGE \\
\hline
\multirow{2}{*}{Full}
&  Pipelined  & 2.09 & \bf 14.4 & \bf 12.4 & \bf 24.8 \\
& Joint  & \bf 1.86 & 12.9 & 11.3 & 22.3\\
\hline
\multirow{2}{*}{\Filtered{}}
& Pipelined & 2.16 & \bf 12.4 & \bf 10.0 & \bf 21.0 \\
& Joint  & \bf 2.03 & 11.4 & 9.2 & 18.7 \\

\hline

\end{tabular}
\end{center}
\vspace{-10pt}
\caption{\small \label{table:combined-overlap} Performance at $t_p$ on examples for which both systems predicted $t_p \leq t_g$ (614 of full and 304 of filtered test sets). All differences are statistically significant.}
\vspace{-10pt}
\end{table*}

\subsection{Evaluation Setup}
\label{sec:combined-human-eval}
We train on filtered data since we found this to improve performance.
At test time, a system can generate a solution description at $t_p \leq t_g$, or it can fail to predict the positive label before or at $t_g$. After a commit/PR for fixing the bug is made at $t_g$, the state of the discussion changes, with possible mentions of the solution that is implemented. Since using this as context to generate a solution description can be considered ``cheating,'' we do not make predictions for time steps after $t_g$. We treat this as the system
\textit{refraining} from generating 
after not finding sufficient context. 

\subsection{Results: Automated Metrics}
The pipelined and joint systems refrained from generating 33.3-35.4\% and 36.4-39.8\% of the time respectively. We present automated metrics for the remaining cases in Table~\ref{table:combined-automatic}. We find that
$t_g-t_p$ is between 1.69 and 1.85 for the pipelined system and between 1.81 and 1.97 for the joint system. While a system should wait until sufficient context is available, sometimes, the last couple utterances before the implementation do not add context about the solution but are personal exchanges
(e.g., ``Thanks'', ``I'll open a PR''). So, generating slightly before $t_{g}$ is acceptable in some cases. Moreover, despite generating early in some cases, the generated output @$t_p$ achieves comparable performance to that @$t_g$, with respect to the generation metrics (BLEU, METEOR, and ROUGE).

Note that the numbers are not directly comparable across the two systems since the exact subset of examples for which $t_p \leq t_g$ varies between the two. In Table~\ref{table:combined-overlap}, we present results for the subset of examples for which both systems predict $t_p \leq t_g$. The joint system achieves lower average error ($t_g - t_p$) for classification while the pipelined system performs better on generation metrics.
 
\subsection{Results: Human Evaluation}
We also do human evaluation, for which we recruited 6 graduate students with 3+ years of Java experience.
Each user evaluated outputs of the two systems for 20 random examples from the filtered test set. Users are given the same information as Section~\ref{sec:HowHumanEval}.
If the system refrained from generating, we ask them if there is sufficient context about the solution at any \TimeUnit{} $t \leq t_g$.
Otherwise, we show them the generated description and ask if there is sufficient context about the solution at $t_p$ and also to rate the informativeness of the description on a Likert scale: 
1: incomprehensible, completely incorrect, irrelevant; 2: generic, rephrasing problem; 3: includes some useful information but does not capture the solution; 4: partially captures solution; 5: completely captures solution.

In the cases that the system generated a description, users found there to be sufficient context at $t_p$ 39.0\% and 33.8\% of the time for the pipelined and joint systems, with average informativeness being 3.3 for both. This suggests that when sufficient context is available, these systems generate descriptions which can be useful for bug resolution.

Because a real-time system must act at a given \TimeUnit{} agnostic to future activity, classifying \textit{when to generate} is challenging. It should defer generation to later time steps if the optimal context is not available. Generating too early can result in output that is generic and re-states the problem. For the cases in which the system generated a description \textit{without} sufficient context at $t_p$, the average informativeness ratings were 2.2 (pipelined) and 2.0 (joint). 
However, deferring generation for too long by expecting more context to emerge later also poses a risk. After the solution has already been implemented, it is too late for a generated description to be useful towards resolving the bug. In the cases that the pipelined and joint systems refrained from generating, there was sufficient context about the solution 34.2\% and 37.0\% of the time respectively. 

Despite the pipelined and joint systems having nuanced differences, we find them to perform similarly. Through our evaluation of these systems, we demonstrate room for improvement, particularly for the classification component in determining the optimal \TimeUnit{} for generation. We leave it to future work to develop more intricate end systems.

\section{Related Work}
\noindent\textbf{Bug report summarization}: To help developers  gather information from bug reports, there is interest in automatic bug report summarization. Approaches for this are designed to generate holistic summaries of bug reports, with a summary being 25\% of the length of the bug report~\cite{LiuBugSum}. We instead aim to generate a concise description that captures a specific aspect of the bug report. Next, bug report summaries are not widely available, so approaches for this task rely on unsupervised techniques~\cite{LiUnSupBugSum,LiuBugSum} or supervision from a small amount of data~\cite{RastkarSum,Jiang2016}. Our approach for obtaining noisy supervision allows us to train supervised models on a large amount of data. Bug report summarization is a post hoc task, done after the bug has been resolved, to help developers address related bug reports in the future. In contrast, our goal is to help resolve the present bug report, so our system must learn \textit{when} to perform generation during an ongoing discussion. Approaches for bug report summarization have been predominantly extractive whereas ours is abstractive.
While we are interested in how bug report summarization techniques fair on our task, their implementations are not publicly available.

\noindent\textbf{Commit message generation}: Unlike the task of automatically generating commit messages to describe code changes that have already been made~\cite{loyola-etal-2017-neural, XuCommit}, our system aims to generate natural language descriptions that can drive code changes. 

\noindent\textbf{Response triggering}: Classifying when to generate a description relates to chatbots learning to respond at an appropriate time~\cite{LiuCustomer} in dyadic conversations. The goal is to avoid interrupting a user who splits up an utterance across multiple turns. We instead consider multi-party dialogue in which an agent should wait until a specific type of content emerges in the discussion. 
\citet{bohus-horvitz-2011-multiparty} studied turn-taking decisions in spoken dialogue systems, using audio-visual features, while ours is a text-based system.

\noindent\textbf{Dialogue + software}: We view our work as a step towards building a dialogue agent for streamlining software bug resolution. There has been minimal work in building interactive systems for this domain, with the exception of a few for tasks like query refinement~\cite{ZhangChatbot} and code generation~\cite{chaurasia-mooney-2017-dialog,yao-etal-2019-model}. \citet{WoodSpeechActs} recently built a dialogue corpus through a ``Wizard of Oz" experiment to study the potential of a Q\&A assistant during bug fixing. \citet{lowe-etal-2015-ubuntu} developed a dialogue corpus based on Ubuntu chat logs to study Q\&A assistants for technical support. In contrast, our dataset is designed for building a collaborative agent that participates in multi-party conversations rather than one which answers directed questions.

\section{Conclusion}
We presented the novel task of generating concise natural language solution descriptions to guide developers in absorbing information relevant towards bug resolution from long discussions. We established benchmarks for this task using a dataset that we constructed with supervision derived from commit messages and pull request titles. Through automated and human evaluation, we demonstrated the utility of these models and also highlight their shortcomings, to encourage more research in exploring ways to address these challenges. We also simulated a real-time setting through two approaches for combining a generation model with a classification component for determining when sufficient context for generating an informative description emerges in an ongoing discussion. We believe this lays the groundwork for future work on building a dialogue agent that participates in bug report discussions to foster efficient resolution.

\section*{Acknowledgements}
We would like to thank Tanya Goyal, Prasoon Goyal, Adrian Benton, and Eunsol Choi for early feedback on this work. We would also like to thank reviewers for their detailed comments and suggestions. This work was supported by NSF grant IIS-1850153, the Bloomberg Data Science Fellowship and a Google Faculty Research Award.

\section*{Ethics Statement}
Our work aims to expedite bug resolution by mobilizing developers and guiding them in absorbing content in long discussions that is relevant towards implementing the solution. Through this, we hope to reduce the life span of software bugs and vulnerabilities that can significantly disrupt everyday operations. Our system is designed to \textit{assist} developers and should not be considered as a replacement for the critical reasoning that is needed during bug resolution. Over-relying on this system to always alert developers when a solution has been recommended could have the opposite effect of causing delays in bug resolution for cases that the system is unable to handle. Additionally, if developers choose to rely solely on the system's generated description and ignore the discussion context, the solutions they implement could potentially be incomplete or incorrect, if the system's output misses important details. Instead, developers should use the generated output to guide their focus and understanding as they read through the discussion.

To build our system, we used data from GitHub, in accordance with its acceptable use policy, and no additional permission was required. Namely, the policy states: ``Researchers may use public, non-personal information from the Service for research purposes, only if any publications resulting from that research are open access.''\footnote{\url{https://docs.github.com/en/github/site-policy/github-acceptable-use-policies}} We use only publicly available data and use it only for research purposes. Additionally, the data we used to train and evaluate models (and publicly release) does not contain personal information (e.g., usernames of users who authored utterances and linked mentions). We require that any future work using our dataset must abide by GitHub's official policy as well.
For evaluation, we conducted human evaluation, for which participants willfully volunteered to be part of the study. They were not compensated for their participation.

\bibliography{custom}
\bibliographystyle{acl_natbib}

\clearpage
\appendix

\section{Data Cleaning}
\label{sec:data-cleaning}

We focus on closed bug reports from the top 1,000 Java projects (in terms of number of stars), as a way of identifying well-maintained projects~\cite{Jarczyk2014GitHubPQ}. We require there to be at least two distinct ``actors" in the discussion, in which the actor can either be a developer who makes an utterance in the discussion or an actor who implements the solution through a commit or pull request. We discard examples in which the reference description is identical to the title (disregarding stopwords), as these are cases in which either the reference description only states the problem and is uninformative or the title already puts forth a solution (in which case a generated description would not be useful). We remove examples with commits or pull requests which simultaneously address multiple bug reports.

We mined 141,389 issues (from 770 of the top 1,000 projects). After applying heuristics, we get 35,010 (from 525 projects), which will be released. Of these, 16,899 pertain to bugs and 18,111 pertain to non-bugs. From the 16,899 bug-related issues, we focus on the 12,328 issues with a single commit message/PR title. We explain our reasoning for discarding examples linked to multiple commits and/or pull requests in Section~\ref{sub:data_collection}. However, such examples (which are available in the data we release) can be useful for supporting generating descriptions at multiple time steps in future work.

From an example's description, we remove references to issue and pull request numbers, as they do not contribute to the meaning and are instead used as identifiers for organizational purposes.

\begin{table*}[t]
\begin{center}
\small
\begin{tabular}{llrrrrr}
\hline
& \bf Model & \BLEU & \METEOR & \bf ROUGE-1 & \bf ROUGE-2 & \bf ROUGE-L \\
\hline
\multirow{17}{*}{Full}
& \ExSumm{} & 0.537 & 0.536 & 0.807 & 0.010 & 0.767 \\
& LexRank & 2.252 & 1.851 & 2.629 & 0.061 & 2.470\\
& \FirstUtteranceOne{} & 4.793 & 6.537 & 10.077 & 2.534 & 8.752 \\
& \FirstUtteranceThree{} & 3.085 & 7.955  & 9.778 & 2.303 & 8.687 \\
& $U_{t_g}$ & 2.842 & 5.425 & 7.426 & 1.363 & 6.712  \\
& \LastUtteranceOne{} & 4.028 & 4.453 & 7.736 & 1.451 & 6.889 \\
& \LastUtteranceThree{} & 3.189 & 5.692 & 8.153 & 1.504 & 7.359 \\
& $U_{t_{g}}$ (Last sentence) & 3.475 & 3.480 & 6.089 & 0.930 & 5.476 \\
& $U_{t_{g}}$ (Last 3 sentences) & 3.234 & 5.082 & 7.525 & 1.287 & 6.787\\
& \RetTitleTitle{} & 6.866 & 4.497 & 11.517 & 1.281 & 10.748 \\
& \RetTitleDescription{} & 8.763 & 6.167 & 15.965 & 2.426 & 14.776 \\
& \ProjectRetTitleTitle{} & 7.442 & 4.709 & 11.501 & 1.49 & 10.943 \\
& \ProjectRetTitleDescription{} & 9.118 & 6.299 & 14.949 & 2.232 & 14.089 \\
\cline{2-7}
& \CopyTitle{} & 14.358 & 13.142 & 27.361 & 11.539 & 24.427 \\
& \SeqToSeq{} & 12.583 & 9.838 & 27.589 & 4.258 & 25.024 \\
& \HRED{} & 12.365 & 9.564  & 26.785 & 3.672 & 24.084 \\
& PLBART & \bf 16.551 & \bf 14.484 & \bf 31.564 & \bf 11.549 & \bf 28.295 \\
& \PLBARTFiltered{} & 14.188 & 12.302 & 27.443 & 8.349 & 25.128 \\

\hline
\multirow{17}{*}{\Filtered{}}
& \ExSumm{} & 0.711 & 0.653 & 1.084 & 0.005 & 1.029  \\
& LexRank & 2.442 & 1.946 & 2.843 & 0.066 & 2.637 \\
& \FirstUtteranceOne{} & 4.951 & 6.207 & 9.881 & 1.938 & 8.553 \\
& \FirstUtteranceThree{} & 3.055 & 7.907 & 9.890 & 1.875 & 8.777  \\
& $U_{t_g}$ & 2.899 & 6.045 & 8.081 & 1.507 & 7.346  \\
& \LastUtteranceOne{} & 4.406 & 4.808 & 8.424 & 1.507 & 7.590 \\
& \LastUtteranceThree{} & 3.356 & 6.257 & 8.894 & 1.681 & 8.060 \\
& $U_{t_{g}}$ (Last sentence) & 3.515 & 3.961 & 6.547 & 1.046 & 5.868 \\
& $U_{t_{g}}$ (Last 3 sentences) & 3.345 & 5.722 & 8.200 & 1.460 & 7.448 \\
& \RetTitleTitle{} & 6.117 & 3.727 & 9.546 & 0.711 & 8.965 \\
& \RetTitleDescription{} & 6.998 & 4.542 & 12.082 & 1.257 & 11.410 \\
& \ProjectRetTitleTitle{} & 6.646 & 4.195 & 9.603 & 1.273 & 9.255 \\
& \ProjectRetTitleDescription{} & 7.593 & 5.064 & 11.895 & 1.638 & 11.328 \\
\cline{2-7}
& \CopyTitle{} & 9.962 & 8.291 & 18.538 & 4.943 & 16.641 \\
& \SeqToSeq{} & 10.168 & 7.521 & 21.846 & 2.278 & 20.116 \\
& \HRED{} & 9.893 & 7.369 & 21.562 & 2.131 & 19.649  \\
& PLBART & \bf 12.319 & 9.877 & 23.419 & 5.452 & 21.097 \\
& \PLBARTFiltered{} & 12.266 & \bf 10.218 & \bf 23.786 & \bf 5.712 & \bf 21.857 \\

\hline

\end{tabular}
\end{center}
\vspace{-10pt}
\caption{\small \label{table:supp-gen-results-table} Comparing models in main paper with  low-performing baselines for generating solution descriptions. Scores for \ExSumm{} are averaged across three trials.}
\vspace{-10pt}
\end{table*}

\section{Details of \HRED{} Model}
\label{app:hred-details}
We encode $U_t$ using a transformer-based encoder and feed the contextualized representation of its first token (\UtteranceStart{}) into the RNN-based discussion encoder to update the \textit{discussion state}, $s_t$. When encoding $U_t$, we also concatenate $s_{t-1}$ to embeddings, to help the model relate $U_t$ with the broader context of the discussion. Note that we treat the title as $U_0$ in the discussion. This process continues until $U_{t_g}$ is encoded, at which point all accumulated token-level hidden states are fed into a transformer-based decoder to generate the output.

Unlike the \SeqToSeq{} model which is designed to reason about the full input at once, this approach reasons step-by-step, with self-attention in the utterance encoder only being applied to tokens within the same utterance. Since the input context for this task is often very large, we investigate whether it is useful to break down the encoding process in this way. We also equip this model with a pointer generator network.

\section{Additional Generation Baselines}
\label{app:gen-baselines}

\begin{table*}[t]
\begin{center}
\small
\begin{tabular}{lllllll}
\hline
& \bf Model & \BLEU & \METEOR & \bf ROUGE-1 & \bf ROUGE-2 & ROUGE-L \\
\hline
\multirow{4}{*}{Full}
& mBART base (randomly initialized) & 9.978 & 6.976 & 17.000 & 2.498 & 15.744 \\
& mBART large  & 15.251 & 12.503 & 28.522 & 9.520 & 26.109\\
& BART base & 14.226 & 11.522 & 26.957 & 8.864 & 24.746 \\
& PLBART & \bf 16.551 & \bf 14.484 & \bf 31.564 & \bf 11.549 & \bf 28.295 \\

\hline
\multirow{5}{*}{\Filtered{}}
& mBART base (randomly initialized) & 8.819 & 6.151 & 14.870 & 2.011 & 13.574\\
& mBART large & 11.663 & 9.233 & 22.295 & 5.159$\ssymbol{2}$ & 20.458\\
& BART base & 10.820 & 8.583 & 21.247 & 5.055$\ssymbol{2}$ & 19.537  \\
& PLBART & \bf 12.319 & \bf 9.877 & \bf 23.419 & \bf 5.452 & \bf 21.097 \\

\hline

\end{tabular}
\end{center}
\vspace{-10pt}
\caption{\small \label{table:bart}Comparing performance of BART-based models. Training/fine-tuning is done with our full training set. Differences that are \textit{not} statistically significant are shown with matching symbols.}
\vspace{-10pt}
\end{table*}

We considered additional baselines; however, since they were performing much lower than other approaches (on wide statistically significant margins), we chose to exclude them from the main paper. We briefly describe these baselines below.

\subsection{Extractive Baselines}
\noindent\textbf{\ExSumm{}}: Using a greedy approach for obtaining noisy extractive summaries~\cite{NallapatiSumma}, we train a supervised extractive summarization model, similar to~\cite{liu-lapata-2019-text}. 

\noindent\textbf{LexRank}: We use LexRank~\cite{ErkanLexRank}, an unsupervised graph-based extractive summarization approach. We extract 1 sentence with threshold 0.1.

\noindent\textbf{$\mathbf{U_1}$ (Lead 1)}: This entails simply taking the first sentence of the first utterance, intended to simulate the Lead-1 baseline that is commonly used in summarization.

\noindent\textbf{$\mathbf{U_1}$ (Lead 3)}: This entails simply taking the first 3 sentences of the first utterance, intended to simulate the Lead-3 baseline that is commonly used in summarization.

\noindent\textbf{$\mathbf{U_{t_g}}$}: Since some part of the solution is often mentioned within $U_{t_{g}}$, we copy this utterance.

\noindent\textbf{$\mathbf{U_{t_g}}$ (Lead 1)}: Since the length of an utterance is quite different than that of a description (Table~\ref{table:data-table}), we extract only the lead sentence of $U_{t_{g}}$.

\noindent\textbf{$\mathbf{U_{t_g}}$ (Lead 3)}: For the reason stated above, we also apply the Lead-3 baseline to this utterance.

\noindent\textbf{$\mathbf{U_{t_g}}$ (Last sentence)}: Rather than extracting the lead sentence, we extract the last sentence of $U_{t_{g}}$.

\noindent\textbf{$\mathbf{U_{t_g}}$ (Last 3 sentences)}: Rather than extracting the lead 3 sentences, we try extracting the last 3 sentences of $U_{t_{g}}$.

\subsection{Retrieval Baselines}
\noindent\textbf{\RetTitleTitle{}}: Using TF-IDF, we compute cosine similarity between the test example's title and titles in the training set, to identify the closest training example, from which we take the description.

\noindent\textbf{\RetTitleDescription{}}: Using TF-IDF, we compute cosine similarity between the test example's title and \textit{descriptions} in the training set, to identify the closest training example, from which we take the description.

\noindent\textbf{\ProjectRetTitleTitle{}}: Using TF-IDF, we compute cosine similarity between the test example's title and titles \textit{for the same project} in the training set, to identify the closest training example, from which we take the description.

\noindent\textbf{\ProjectRetTitleDescription{}}: Using TF-IDF, we compute cosine similarity between the test example's title and descriptions for the same project in the training set, to identify the closest training example, from which we take the description.

\subsection{Baseline Results}
We present baseline results in Table~\ref{table:supp-gen-results-table}. In addition to the metrics used in the main paper, we report ROUGE-1 and ROUGE-2. All of these baselines substantially underperform models presented in the main paper, especially the \ExSumm{} model. We believe this model performs so poorly due to noise in the supervision and because the extracted summaries are longer and structured differently than the reference descriptions in our dataset. Additionally, there are many examples in which the model does not select a single sentence from the input, resulting in the prediction being the empty string. LexRank also performs poorly in terms of automated metrics against the reference description. This unsupervised approach aims to identify a ``centroid" sentence that summarizes the full input context and is not designed to specifically focus on solution-related context. 

All baselines that extract a whole utterance or sentences from specific utterances perform poorly, demonstrating the need for content selection from the broader context and content synthesis rather than relying on simple heuristics to produce a description of the solution. We find that the retrieval baselines tend to achieve higher scores, as retrieved descriptions are from the same distribution as the reference descriptions. However, these numbers are still much lower than those in the main paper.

\section{BART Models}
\label{app:bart}

We use PLBART~\cite{AhmadPLBART}, which was pretrained on large amounts of code from GitHub and software-related natural language from StackOverflow. Compared to other pretrained models, fine-tuning PLBART achieves higher performance for various NL+code tasks, including code summarization, code generation, code translation, and code classification. Since our task also requires reasoning about code and technical text, we choose PLBART over other pretrained models in our work. We present automated metrics for PLBART and \PLBARTFiltered{} in Table~\ref{table:gen-results-table}. The average length of PLBART's output is 9.0 and 8.6 tokens on the full and filtered test sets respectively, while it is 9.3 and 9.4 for \PLBARTFiltered{}.

For completion, we compare against BART-based models which are not pretrained on code or technical text. First, we consider mBART base (multilingual BART)~\cite{tang2020multilingual}, which is the underlying architecture of PLBART. Without pretraining (randomly initializing the same architecture), performance is very low, as shown in Table~\ref{table:bart}. The publicly released pretrained mBART model, which is pretrained on non-technical natural language, does not use the base architecture but rather large. We also fine-tune this model on our training set but find that it achieves lower performance than PLBART. Finally, we compare against BART base~\cite{lewis-etal-2020-bart}, which is also pretrained on non-technical natural language. Again, this model underperforms PLBART. Because PLBART's performance is higher, we choose to focus on this model in our work.

\section{Human Evaluation Setup}
In the user study, users are shown the title of the bug report, all utterances up till (and including) $U_{t_{g}}$, and the reference description in our dataset for the given example. We choose to provide this as a manual suggestion to help guide users in better understanding a bug report, for a software project with which they have minimal familiarity. However, we state in our instructions that this is merely provided for reference and is not necessarily the exact and only valid answer.

Next, we show them up to 5 model predictions and ask them to ``select the one(s) which add(s) the most amount of useful information that will help resolve the bug, beyond just re-stating the problem itself." Note that these are presented in random order (per example), without any identifying information about the underlying models that generated them.
We explain that we consider a description to be informative if it provides content that will be useful towards \textit{fixing} the issue, beyond just rephrasing the problem itself. And we encourage users to select candidates based on content that is informative, rather than focusing on exact phrasing. If all candidates appear to be poor (completely unrelated to the resolving the bug, uninformative, incomprehensible, or plain wrong), users are asked to select another option: ``All candidates are poor." If there is no useful information towards resolving the bug in the context and they are unable to evaluate candidate descriptions, they are asked to select another option: ``The context does not have any useful information for resolving the bug." They must also justify their selection by writing a brief rationale.

This is a challenging task, as it requires reading through and reasoning about a large amount of text to evaluate each example. To prepare annotators, we first present a set of training examples and a training video in which we demonstrate how the task should be completed.

\section{Analyzing \Curated{} Subset}
\label{app:context_usage}
\begin{figure*}
\begin{subfigure}{.49\textwidth}
  \centering
  \includegraphics[width=\columnwidth]{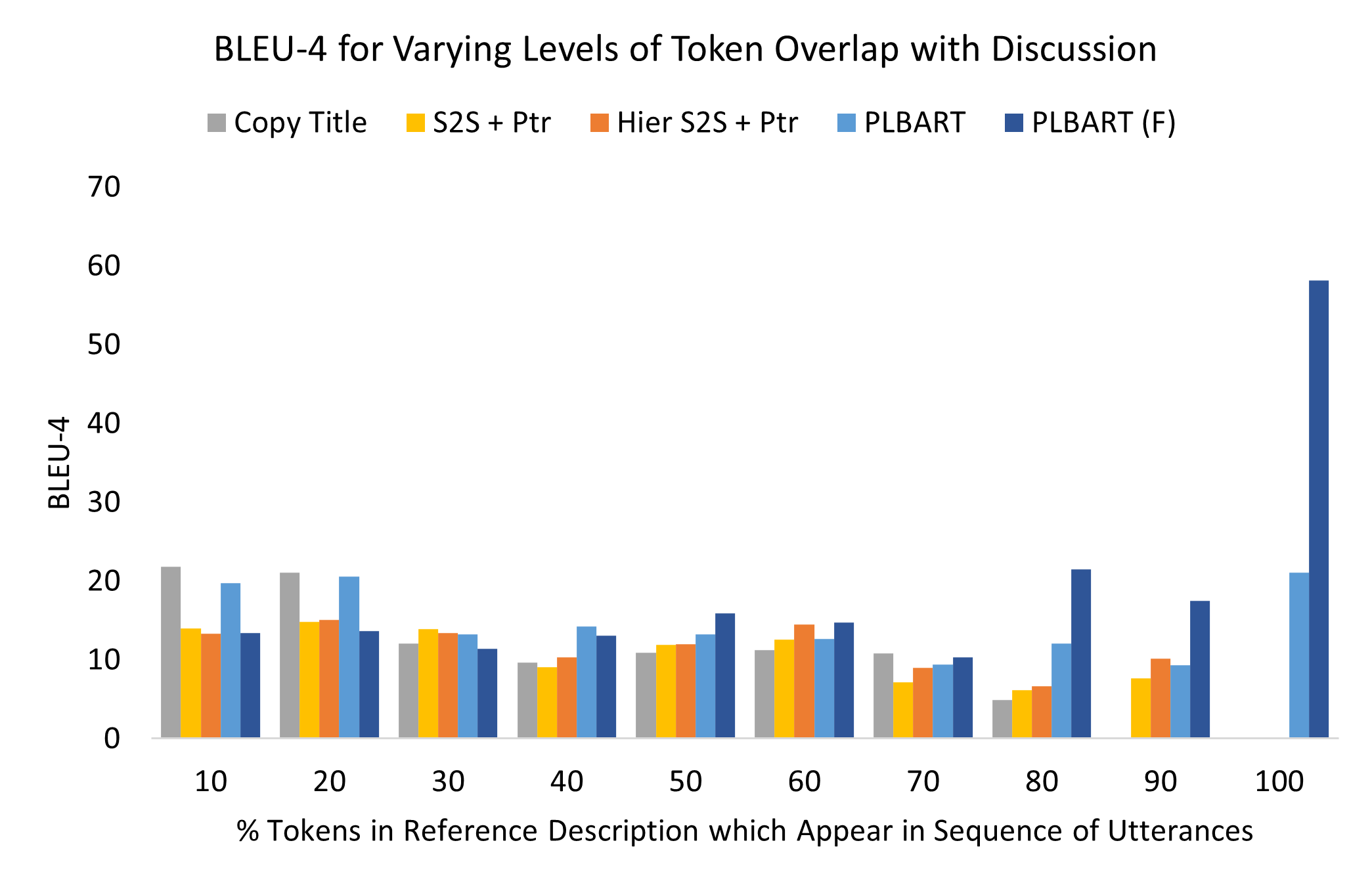}
  \caption{BLEU-4}
  \label{fig:bleu_4_plot}
\end{subfigure}%
\begin{subfigure}{.49\textwidth}
  \centering
  \includegraphics[width=\columnwidth]{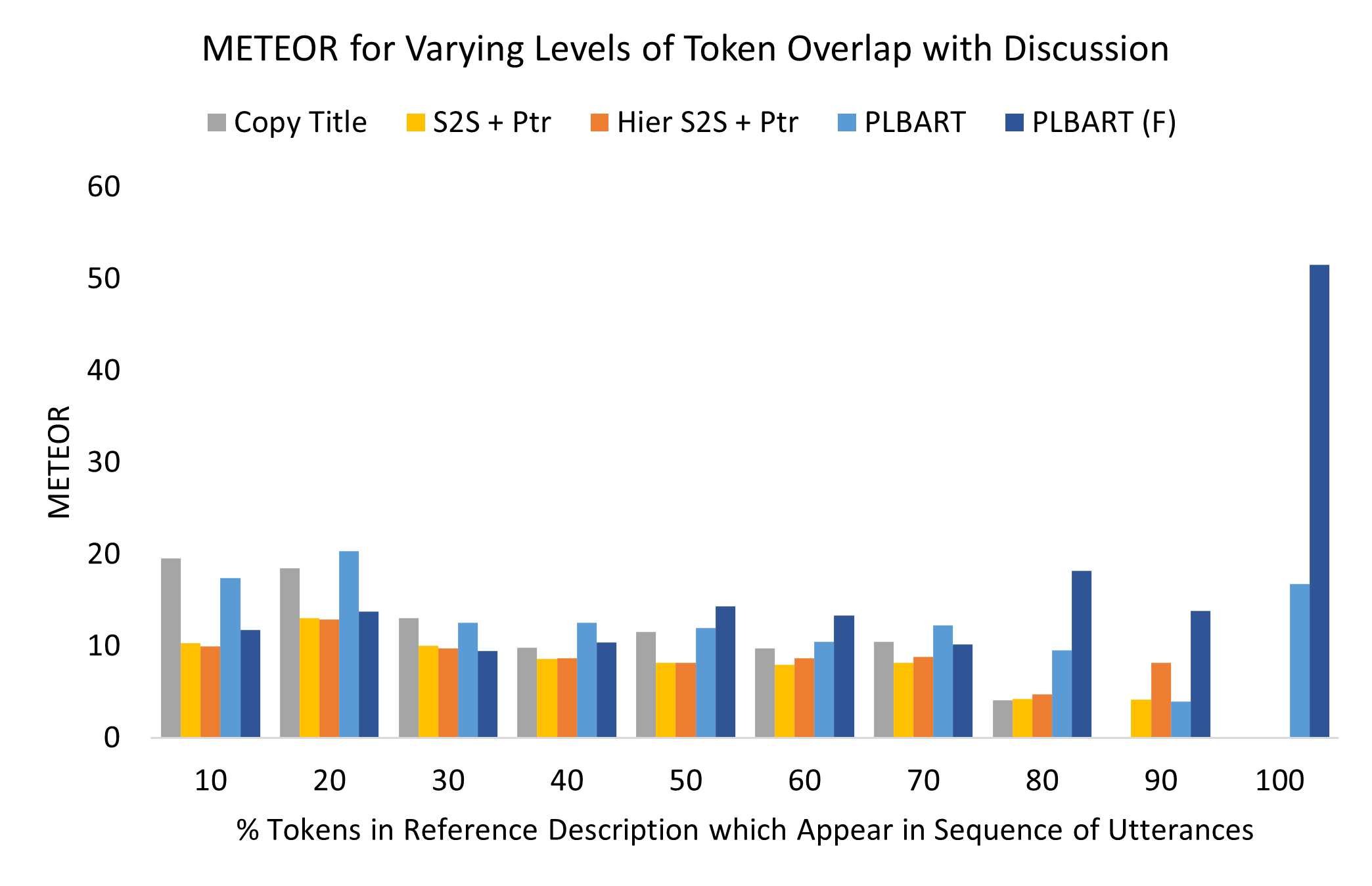}
  \caption{METEOR}
  \label{fig:meteor_plot}
\end{subfigure}
\begin{subfigure}{.49\textwidth}
  \centering
  \includegraphics[width=\columnwidth]{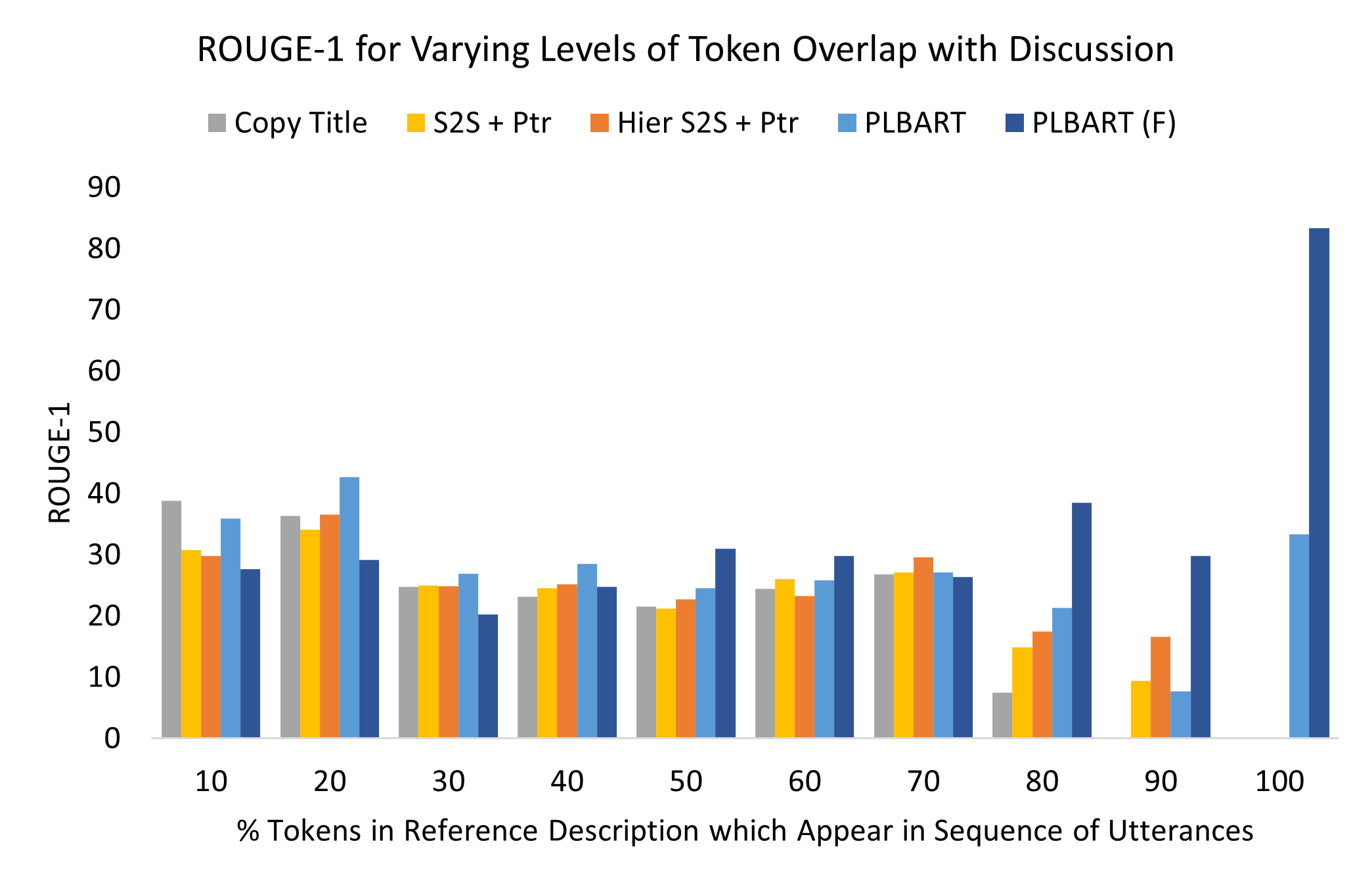}
  \caption{ROUGE-1}
  \label{fig:rouge_1_plot}
\end{subfigure}
\begin{subfigure}{.49\textwidth}
  \centering
  \includegraphics[width=\columnwidth]{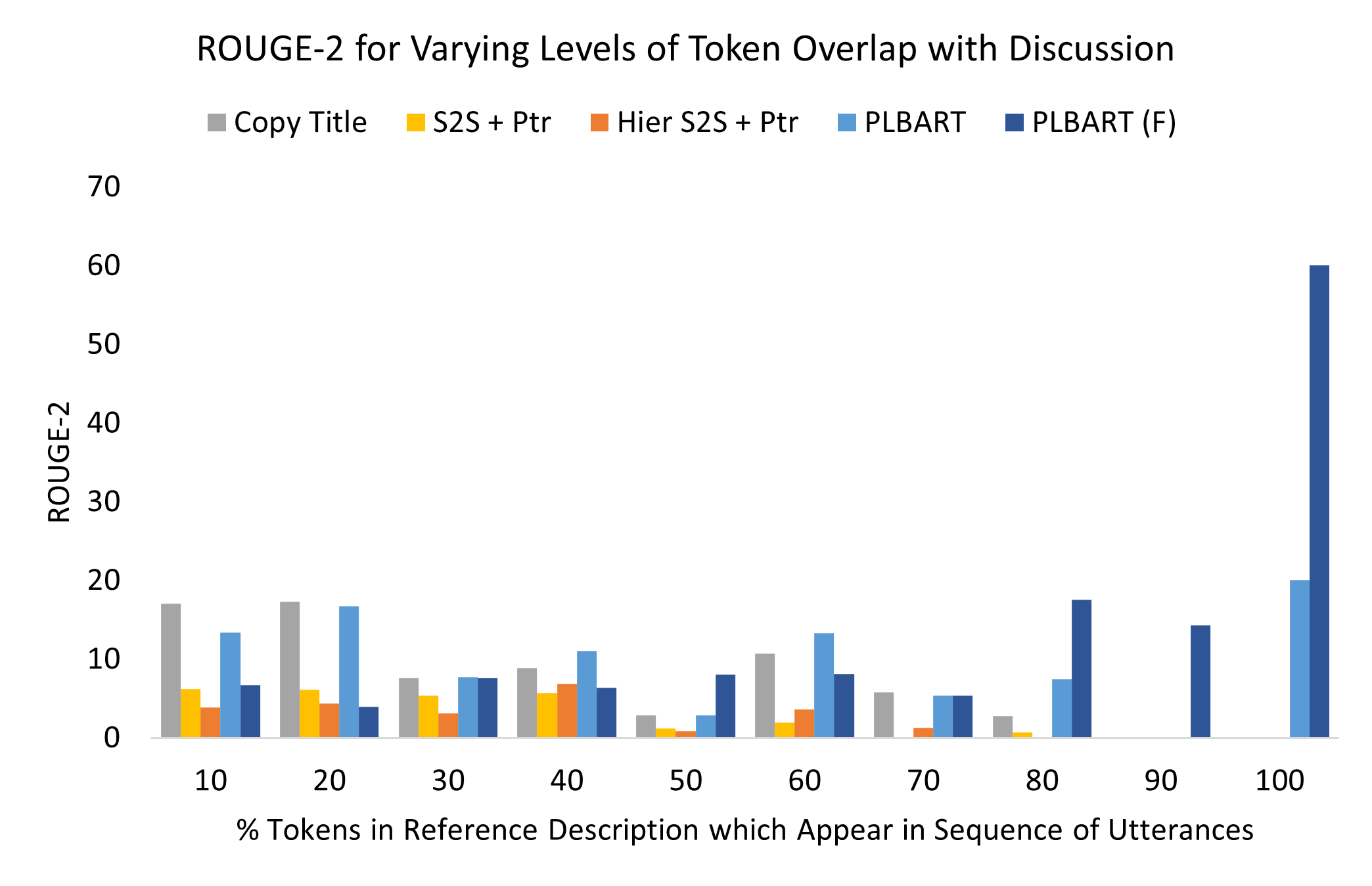}
  \caption{ROUGE-2}
  \label{fig:rouge_2_plot}
\end{subfigure}
\begin{subfigure}{.49\textwidth}
  \centering
  \includegraphics[width=\columnwidth]{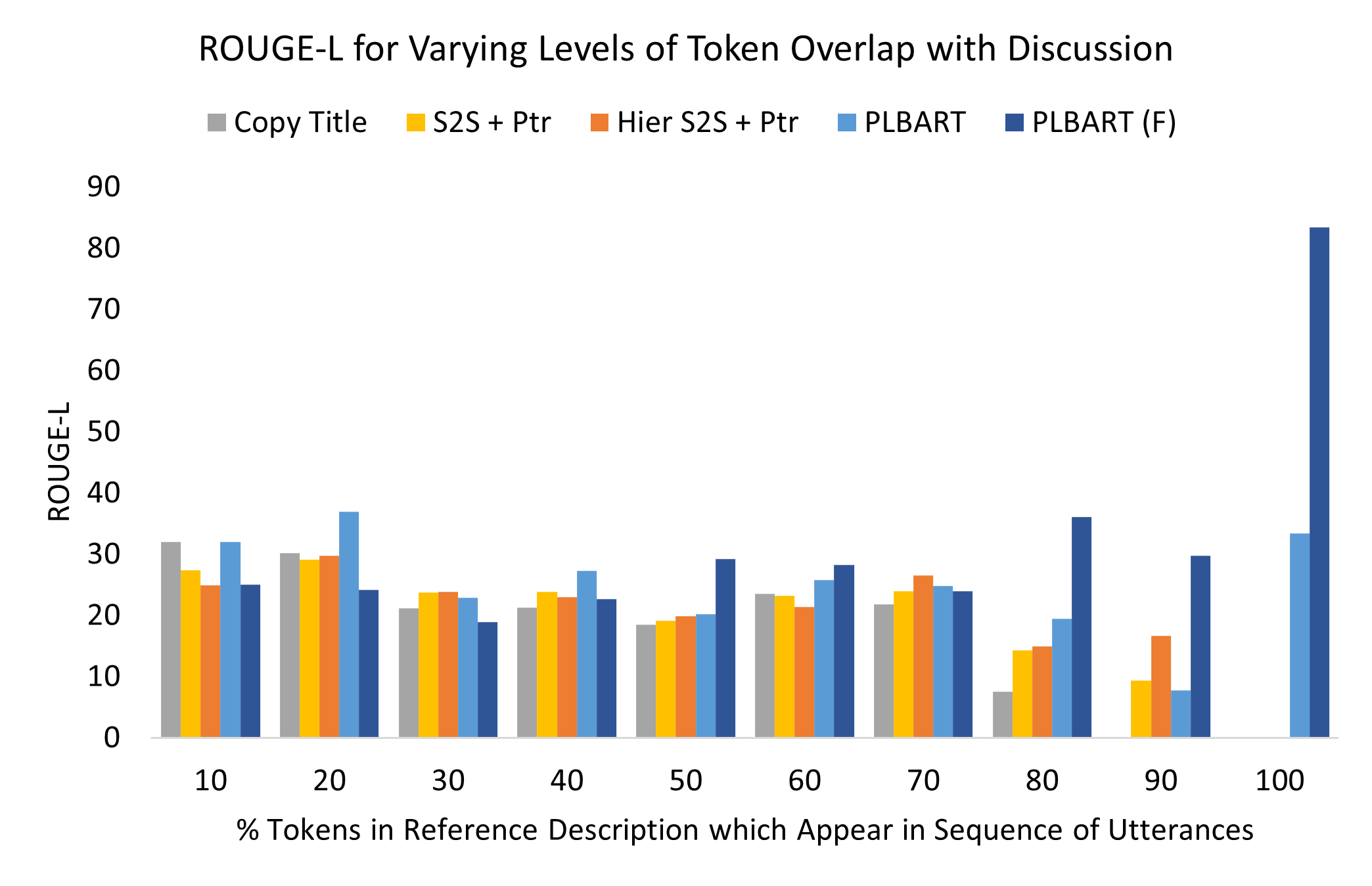}
  \caption{ROUGE-L}
  \label{fig:rouge_l_plot}
\end{subfigure}%
\vspace{-10pt}
\caption{Metrics for \Curated{} subset, with buckets corresponding to the \% of tokens in reference description which also appear in $U_1...U_{t_g}$ (disregarding title tokens). Bucket 10 corresponds to [0, 10)\%, 20 corresponds to [10, 20)\%, etc.}
\vspace{-10pt}
\label{fig:plots}
\end{figure*}

The \Curated{} subset consists of 109 examples from the test set spanning 45 projects, with average $T=4.1$ and $t_g = 3.2$. We present automated metrics for this subset in Table~\ref{table:curated}.
Results are analogous to the full test set, except that the numbers are generally lower for all models other than for \PLBARTFiltered{}, which achieves consistent performance. \PLBARTFiltered{} slightly underperforms PLBART on automated metrics overall. However, this is because these metrics are computed against the single reference description, which could diverge from how the solution is formulated in the discussion since the developer could have written an uninformative/generic description. To do more fine-grained analysis, in Figure~\ref{fig:plots}, we plot automated metrics for varying percentages of token overlap between the reference description and $U_1...U_{t_g}$ (excluding tokens already present in the title which have been used to state the problem). Higher overlap suggests that the reference description draws more content from within the discussion. For higher percentages, \PLBARTFiltered{} generally achieves higher scores against the reference than PLBART and all other models, indicating that this model is better at gathering information from within the discussion. In Table~\ref{table:ngrams}, we supplement the n-gram analysis from Section~\ref{sec:curated_analysis}.

\begin{table}[t]
\begin{center}
\small
\begin{tabular}{llll}
\hline
\bf Model & \BLEU & \METEOR &  \ROUGEL  \\
\hline
\CopyTitle{} & 12.6 & 12.2$\ssymbol{5}$ & 22.1 \\
\SeqToSeq{} &  11.6 & 8.9  & 23.1 \\
\HRED{} & 12.0 & 9.0 & 22.9 \\
PLBART & \bf 14.6 & \bf 13.2 & \bf 26.0  \\
\PLBARTFiltered{} & 14.2 & 12.3$\ssymbol{5}$ & 25.1 \\

\hline

\end{tabular}
\end{center}
\vspace{-10pt}
\caption{\small \label{table:curated}Automated metrics for generation on \Curated{} subset. Differences that are \textit{not} statistically significant are indicated with matching symbols.}
\vspace{-10pt}
\end{table}

\begin{table*}[t]
\begin{center}
\small
\begin{tabular}{lllllllllll}
\hline
& &  \multicolumn{4}{c}{\bf Title $\downarrow$} && \multicolumn{4}{c}{\bf $U_1...U_{t_g}$ only $\uparrow$}  \\
\cline{3-6}
\cline{8-11}
& \bf  Model &  \bf 1 &  \bf 2 &  \bf 3 &  \bf 4 &&  \bf 1 &  \bf 2 &  \bf 3 &  \bf 4 \\
\hline
\multirow{6}{*}{Full}
& \CopyTitle{} & 100.0 & 100.0 & 100.0 & 100.0 && 0.0 & 0.0 & 0.0 & 0.0\\
& \SeqToSeq{} & 65.6 & 34.4 & 39.3 & 46.5 && 28.6 & 24.9 & 27.0 & 25.0\\
& \HRED{} & 60.2 & 33.9 & 41.1 & 50.4 && 37.4 & 27.9 & 28.3 & 29.2\\
& PLBART & 79.3 & 75.0 & 72.5 & 71.7 && 30.7 & 34.8 & 34.6 &  \bf 39.9\\
& \PLBARTFiltered{} & 43.2 & 37.4 & 38.3 & 43.1 &&  \bf 47.1 &  \bf 38.1 &  \bf 35.6 & 37.2 \\
& Reference &  \bf 35.1 &  \bf 30.9 &  \bf 33.5 &  \bf 37.7 && 34.5 & 22.2 & 22.2 & 25.3\\

\hline
\multirow{6}{*}{Filtered}
& \CopyTitle{} & 100.0 & 100.0 & 100.0 & 100.0 && 0.0 & 0.0 & 0.0 & 0.0\\
& \SeqToSeq{} & 64.5 & 33.8 & 39.1 & 38.3 && 29.4 & 25.3 & 23.8 & 0.0\\
& \HRED{} & 58.4 & 33.3 & 39.3 & 45.7 && 40.4 & 28.4 & 30.0 & 29.2\\
& PLBART & 76.9 & 73.4 & 71.1 & 70.4 && 34.0 & 37.0 & 36.3 &  \bf 41.2\\
& \PLBARTFiltered{} & 38.4 & 33.9 & 35.2 & 40.7 &&  \bf 51.0 &  \bf 40.0 &  \bf 36.6 & 38.1 \\
& Reference &  \bf 23.7 &  \bf 18.6 &  \bf 18.4 &  \bf 16.3 && 40.1 & 22.8 & 21.4 & 23.0\\

\hline

\multirow{6}{*}{\CuratedCap{}}
& \CopyTitle{} & 100.0 & 100.0 & 100.0 & 100.0 && 0.0 & 0.0 & 0.0 & 0.0\\
& \SeqToSeq{} & 64.8 & 37.1 & 38.5 & 22.5 && 31.6 & 25.3 & 33.1 & 25.0\\
& \HRED{} & 60.3 & 34.2 & 37.9 & 28.3 && 38.7 & 26.1 & 29.2 & 0.0\\
& PLBART & 80.8 & 77.7 & 72.8 & 70.3 && 31.0 & 41.4 & 37.0 &  \bf 50.0\\
& \PLBARTFiltered{} & 36.9 & 28.4 & 30.8 &  34.1 &&  \bf 52.8 &  \bf 42.3 &  \bf 39.4 & 45.0\\
& Reference &  \bf 32.7 & \bf 22.2 &  \bf 26.2 & \bf 35.6 && 38.8 & 25.4 & 23.1 & 27.1\\

\hline

\end{tabular}
\end{center}
\vspace{-10pt}
\caption{\label{table:ngrams}Percent of unigrams, bigrams, trigrams and 4-grams in the prediction (or reference) which appear in the title and in $U_1..U_{t_g}$ only (excluding the title). Lower is better for the title and higher is better for $U_1..U_{t_g}$ only.}
\vspace{-10pt}
\end{table*}

\begin{table*}[t]
\begin{center}
\small
\begin{tabular}{llllllll}
\hline
&  & \bf \FirstBaseline{} & \bf \SecondBaseline{} & \bf \RandomBaseline{} & \bf \WeightedRandomBaseline{} & \bf Pipelined & \bf Joint \\
\hline
\multirow{2}{*}{Full}
&  ($\uparrow$) $\mathbf{t_p \leq t_g }$ &  100.0\% & 70.5\% & 76.0\% & 77.1\% & 66.7\% & 60.2\%	\\
& ($\downarrow$) $\mathbf{t_g - t_p}$ & 2.2 & 2.1 & 2.2 & 2.2 & 1.7 & 1.8	\\

\hline
\multirow{2}{*}{\Filtered{}}
& ($\uparrow$) $\mathbf{t_p \leq t_g }$ &  100.0\% & 76.2\% & 79.4\% & 80.1\% & 64.6\% & 63.6\%	\\
& ($\downarrow$) $\mathbf{t_g - t_p}$ & 2.6 & 2.4 & 2.5 & 2.5 & 1.9 & 2.0	\\

\hline

\end{tabular}
\end{center}
\vspace{-10pt}
\caption{\small \label{table:when2speak-results-table}Percent of time $t_{p} \leq t_{g}$ and for these particular cases, the mean absolute error between  $t_g$ and $t_p$.}
\vspace{-10pt}
\end{table*}

\section{Classification Performance}
\label{app:classification}

To benchmark performance on the classification task for determining when sufficient context is available for generating an informative description, we consider some simple baselines. We observe that there are many cases in which $t_{g}=1,2$, i.e., the solution is implemented immediately after the first or second utterance. So, we include  the \FirstBaseline{} baseline which always predicts a positive label at $t=1$, and \SecondBaseline{} which predicts negative at $t=1$ and positive at $t=2$, if $t_{g} \geq 2$ (otherwise it never predicts positive).

We include the \RandomBaseline{} baseline which progresses through the discussion, randomly deciding between the positive and negative label after each utterance, based on a uniform distribution. We also include \WeightedRandomBaseline{}, which instead uses the probability distribution of labels at the example-level estimated from the filtered training set (pos =  $\frac{1}{N}\sum_{n=1}^{N} \frac{1}{t_{g}}$=0.510, neg = 0.490). Results are averaged across 3 trials.
We present results in Table~\ref{table:when2speak-results-table}.

\begin{table*}[h!]
\begin{center}
\small
\begin{tabular}{llllll}
\hline
\bf Model & \bf BLEU-4 & \bf METEOR & \bf ROUGE-1 & \bf ROUGE-2 & \bf ROUGE-L \\
\hline

\CopyTitle{} & 15.223 & 13.645 & 28.088 & 12.322 & 25.341 \\
\SeqToSeq{} & 12.896 & 10.241 & 27.757 & 4.571 & 25.921 \\
\HRED{} & 12.758 & 10.119 & 28.722 & 3.934 & 25.275 \\
PLBART & \bf 16.924 & 1\bf 4.979 & \bf 32.152 & \bf 12.124 & \bf 29.623 \\
\PLBARTFiltered{} & 15.059 & 13.057 & 29.107 & 9.111 & 26.710 \\

\hline

\end{tabular}
\end{center}
\vspace{-10pt}
\caption{\label{table:valid-gen-results-table}Scores for generation @ $t_g$ on the 1,232 examples in the full validation set.}
\vspace{-10pt}
\end{table*}

\begin{table*}[h!]
\begin{center}
\small
\begin{tabular}{llll}
\hline
\bf Model & \bf Train &  \bf Eval &  \bf Epoch \\
\hline
\SeqToSeq{} & 2:56:19 & 0:01:12 & 52.0 \\
\HRED{} & 4:47:34 & 0:01:22 & 51.0 \\
PLBART (fine-tuning) & 0:32:07 & 0:00:25 & 11.0 \\
\PLBARTFiltered{} (fine-tuning) & 0:26:08 & 0:00:28 & 15.0 \\
Pipelined system (classifier only) & 2:12:48 & 0:02:09 & 12.0 \\
Jointly trained combined system & 6:25:01 & 0:15:06 & 22.0 \\
\hline

\end{tabular}
\end{center}
\vspace{-10pt}
\caption{\label{table:running-times-table} Average training time, inference time, and number of epochs. Format for time is H:M:S.}
\vspace{-10pt}
\end{table*}

\section{Reproducibility Checklist}

\subsection{Validation Performance}
We report performances on the full validation set. Results for the generation task are in Table~\ref{table:valid-gen-results-table}.

\section{Hyperparameters}
All neural models were implemented using PyTorch. For \SeqToSeq{} and \HRED{}, we use a batch size of 8, an initial learning rate of 3e-05, and a dropout rate of 0.2. Our transformer models have 4 encoder and decoder layers, 4 heads in multi-head attention, a hidden size of 64, and feedforward hidden size 256. We use Adam as the optimizer and have a learning rate scheduler with gamma 0.95 which decays after an epoch if the validation loss has not improved. We use early stopping with patience 5 during training.

For classification, the classification head consists of a linear layer (dimension 768), followed by a tanh non-linear layer, and a final linear projection layer (dimension 2). When computing cross entropy loss for classification, we weight  the positive and negative labels using the inverse of the class proportion to handle class imbalance (1.70 and 0.71 respectively).
For the joint model, loss for a given example is computed as follows, with $\lambda_1=0.8, \lambda_2=0.2$ (tuned on validation data).
\[ L = \lambda_1L_{gen}(t_g) + \lambda_2\sum_{t=1}^{t=t_g} L_{class}(t) \]

\subsection{Tuning}
For \SeqToSeq{} and \HRED{}, hyperparameters are tuned manually. For batch size, we consider \{8,16,32\}, learning rate \{1e-03, 1e-04, 3e-05\}, dropout \{0.1, 0.2, 0.4, 0.5, 0.6\}, encoder/decoder layers \{2, 4, 6, 8\}, number of heads \{2, 4, 8\}, hidden sizes \{32, 62, 128\}, and feedforward dimensions \{128, 256, 512\}. These hyperparameters are tuned on validation data, using the text generation metrics mentioned in Section~\ref{sec:HowAutoEval} for generation. For tuning, we do not do grid search but rather compare performances between models trained with identical configurations, with the exception of a single parameter. Therefore, the number of hyperparameter tuning runs scales linearly. We ran each configuration once.
For PLBART-based models, we use the same configurations as the scripts released by \citet{AhmadPLBART}.

\subsection{Random Seeds}
For the randomly initialized models, random seeds are set according to the timestamp, and we average results across 3 trials. For \SeqToSeq{}, the seeds were: 1620001129, 1620001158, and 1620004022. For \HRED{}, the seeds were: 1620001125, 1620001159, and 1620004024.

\section{Statistical Significance Testing}
We compute statistical significance using bootstrap tests~\cite{berg-kirkpatrick-etal-2012-empirical} with $p < 0.05$ and 10,000 samples of size 5,000 each.  

\subsection{Running Times}
Table \ref{table:running-times-table} reports average training time, inference time, and \# epochs for the various models considered in this work. The PLBART-based models were trained/fine-tuned on NVIDIA DGX GPUs (32 GB) and all other models were trained and evaluated using on GeForce GTX Titan GPUs (8 GB). All models used single-GPU training.

\end{document}